\documentclass[twoside]{article}

\usepackage{tech_report}

\usepackage[round]{natbib}
\renewcommand{\bibname}{References}
\renewcommand{\bibsection}{\subsubsection*{\bibname}}

\usepackage[english]{babel}

\usepackage{amsthm}

\newtheorem{definition}{Definition}

\usepackage{amsmath,amsfonts,bm}

\def\eqref#1{equation~\ref{#1}}

\def\1{\bm{1}}

\def\mC{{\bm{C}}}

\def\mG{{\bm{G}}}

\DeclareMathAlphabet{\mathsfit}{\encodingdefault}{\sfdefault}{m}{sl}
\SetMathAlphabet{\mathsfit}{bold}{\encodingdefault}{\sfdefault}{bx}{n}

\newcommand{\R}{\mathbb{R}}

\usepackage{graphicx}
\usepackage{hyperref}
\usepackage{url}
\usepackage[norelsize,ruled,vlined]{algorithm2e}

\begin{document}

%

%

\twocolumn[

\aistatstitle{Verification of Neural Networks: \\ Specifying Global Robustness using Generative Models}

\aistatsauthor{Nathana{\"e}l Fijalkow \And Mohit Kumar Gupta}

\aistatsaddress{CNRS, LaBRI, Universit{\'e} de Bordeaux, and \\ 
The Alan Turing Institute, London
\And 
Indian Institute of Technology Bombay
} 
]

\begin{abstract}
The success of neural networks across most machine learning tasks 
and the persistence of adversarial examples 
have made the verification of such models an important quest.
Several techniques have been successfully developed to verify robustness,
and are now able to evaluate neural networks with thousands of nodes.
The main weakness of this approach is in the specification:
robustness is asserted on a validation set consisting of a finite set of examples, \textit{i.e.} locally.

We propose a notion of global robustness based on generative models, 
which asserts the robustness on a very large and representative set of examples.
We show how this can be used for verifying neural networks.
In this paper we experimentally explore the merits of this approach, 
and show how it can be used to construct realistic adversarial examples.
\end{abstract}

\section{Introduction}

We consider the task of certifying the correctness of an image classifier, 
\textit{i.e.} a system taking as input an image and categorising it.
As a main example we will consider the MNIST classification task, which consists in categorising hand-written digits.
Our experimental results are later reproduced for the drop-in dataset Fashion MNIST (\cite{XRV17}).

The usual evaluation procedure consists in setting aside from the dataset a validation set, 
and to report on the success percentage of the image classifier on the validation set.
With this procedure, it is commonly accepted that the MNIST classification task is solved, 
with some convolutional networks achieving above 99.7\% accuracy (see \textit{e.g.}~\cite{CMS12,pmlr-v28-wan13}).
Further results suggest that even the best convolutional networks cannot be considered to be robust,
given the persistence of adversarial examples: 
a small perturbation -- invisible to the human eye -- in images from the dataset is enough to induce misclassification (\cite{SZSBEGF13}).

This is a key motivation for the verification of neural networks: 
can we assert the robustness of a neural network, \textit{i.e.} the absence of adversarial examples?
This question has generated a growing interest in the past years at the crossing of different research communities
(see \textit{e.g.}~\cite{HKWW17,KBDJK17,WZCSHDBD18,GMDTCV18,MGV18,GKPB18,KHIJLLSTWZDK19}), 
with a range of prototype tools achieving impressive results.
The robustness question is formulated as follows: given an image $x$ and $\varepsilon > 0$, 
are all $\varepsilon$-perturbations of $x$ correctly classified?

We point to a weakness of the formalisation: it is \textit{local}, meaning it is asserted for a given image $x$ 
(and then typically checked against a finite set of images).
In this paper, we investigate a \textit{global} approach for specifying the robustness of an image classifier.
Let us start from the ultimate robustness objective, which reads:
\begin{quote}
For every category, for every \textit{real-life image} of this category and for every \textit{perturbation} of this image,
the perturbed image is correctly classified.
\end{quote}
Formalising this raises three questions:
\begin{enumerate}
	\item How do we quantify over \textit{all} real-life images?
	\item What are \textit{perturbed} images?
	\item How do we \textit{effectively} check robustness?
\end{enumerate}

In this work we propose a formalisation based on generative models.
A generative model is a system taking as input a random noise and generating images, 
in other words it represents a probabilistic distribution over images.

Our specification depends on two parameters $(\varepsilon,\delta)$.
Informally, it reads:

\begin{quote}
An image classifier is $(\varepsilon,\delta)$-robust with respect to a generative model 
if the probability that for a noise $x$, all $\varepsilon$-perturbations of $x$ generate correctly classified images 
is at least $1 - \delta$.
\end{quote}
The remainder of the paper presents experiments supporting the claims that the global robustness specification 
has the following important properties.

\medskip\textbf{Global.}
The first question stated above is about quantifying over all images.
The global robustness we propose addresses this point by (implicitly) quantifying over a very large and representative set of images.

\medskip\textbf{Robust.}
The second question is about the notion of perturbed images.
The essence of generative models is to produce images reminiscent of real images (from the dataset); 
hence testing against images given by a generative model includes the very important perturbation aspect present 
in the intuitive definition of correctness.

\medskip\textbf{Effective.}
The third question is about effectivity.
We will explain that global robustness can be effectively evaluated for image classifiers built using neural networks.


\subsection*{Related work}
\cite{XLZHLS18} train generative models for finding adversarial examples, 
and more specifically introduce a different training procedure (based on a new objective function)
whose goal is to produce adversarial examples.
Our approach is different in that we use generative models with the usual training procedure and objective,
which is to produce a wide range of realistic images.

\section{Global Correctness}
This section serves as a technical warm-up for the next one: 
we introduce the notion of \textit{global correctness}, a step towards our main definition of \textit{global robustness}.

We use $\R^d$ for representing images with $||\cdot||$ the infinity norm over $\R^d$, 
and let $C$ be the set of categories, so an image classifier represents a function $\mC : \R^d \to C$.

A generative model represents a distribution over images, and in effect is a 
neural network which takes as input a random noise in the form of a $p$-dimensional vector $x$ and produces an image $\mG(x)$.
Hence it represents a function $\mG : \R^p \to \R^d$.
We typically use a Gaussian distribution for the random noise, written $x \sim \mathcal{N}(0,1)$.

Our first definition is of \textit{global correctness}, 
it relies on a first \textit{key} but simple idea, which is to compose a generative model $\mG$ with an image classifier $\mC$:
we construct a new neural network $\mC \circ \mG$ by simply rewiring the output of $\mG$ to the input of $\mC$,
so $\mC \circ \mG$ represents a distribution over categories.
Indeed, it takes as input a random noise and outputs a category.

\begin{figure}[ht]
\centering
\includegraphics[scale=.2]{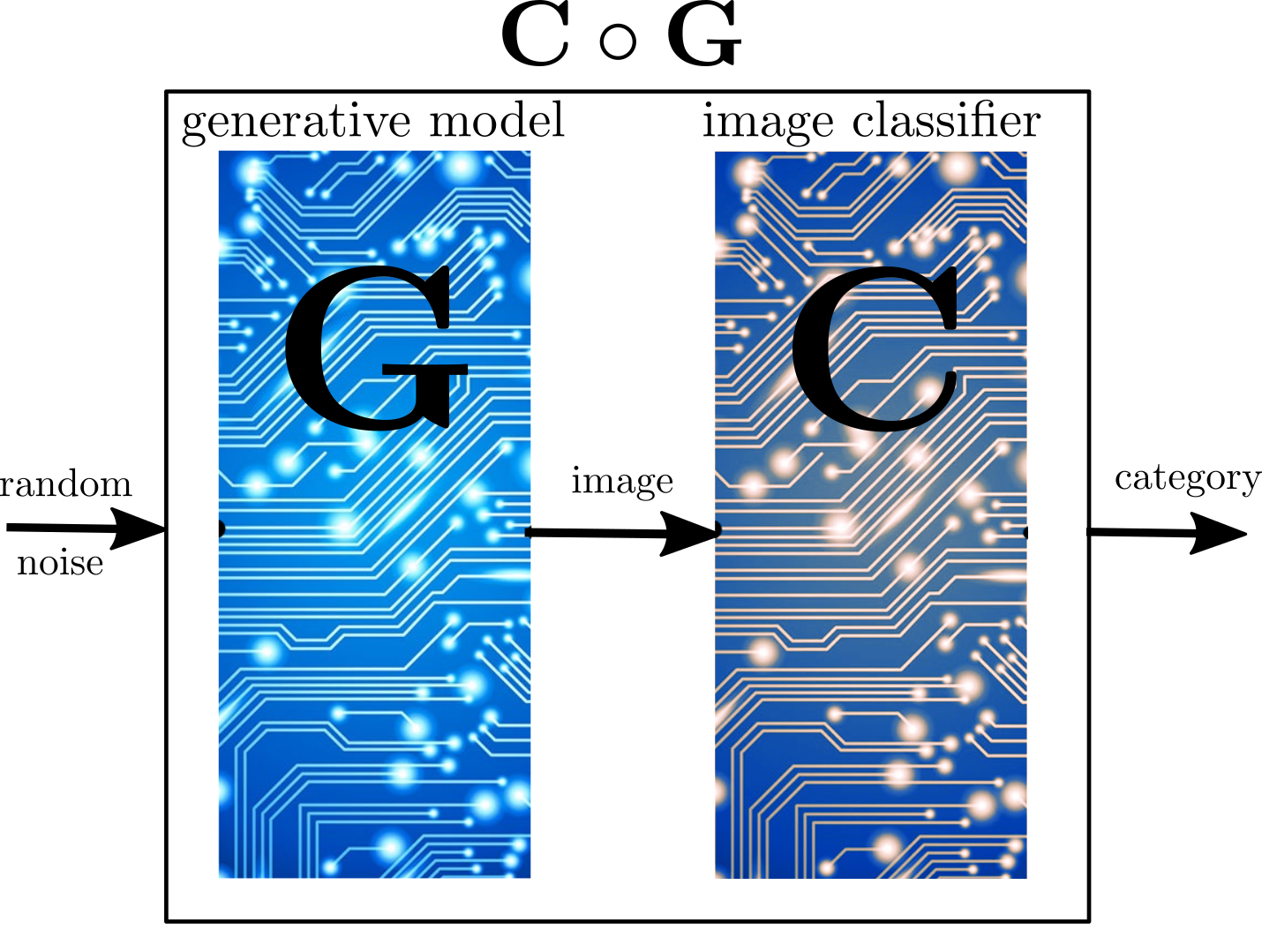} 
\caption{Composition of a generative model with an image classifier}
\label{fig:composition}
\end{figure}

\begin{definition}[Global Correctness]
Given for each $c \in C$ a generative model $\mG_c$ for images of category~$c$,
we say that the image classifier $\mC$ is $\delta$-correct with respect to the generative models $(\mG_c)_{c \in C}$ 
if for each $c \in C$,
\[
\mathbb{P}_{x \sim \mathcal{N}(0,1)}(\mC \circ \mG_c(x) = c) \ge 1 - \delta.
\]
In words, the probability that for a noise $x$ 
the image generated (using $\mG_c$) is correctly classified (by $\mC$) is at least $1 - \delta$.
\end{definition}


\subsection*{Assumptions}
Our definition of global correctness hinges on two properties of generative models:
\begin{enumerate}
	\item generative models produce a wide variety of images,
	\item generative models produce (almost only) realistic images.
\end{enumerate}
The first assumption is the reason for the success of generative adversarial networks (GAN) (\cite{GPMXWOCB14}).
We refer for instance to~\cite{KLA18} and to the attached website~\url{thispersondoesnotexist.com} for a demo.

In our experiments the generative models we used are out of the shelf generative adversarial networks (GAN) (\cite{GPMXWOCB14}),
with $4$ hidden layers of respectively $256$, $512, 1024$, and $784$ nodes, producing images of single digits.

To test the second assumption we performed a first experiment called the \textit{manual score experiment}.
We picked $100$ digit images using a generative model and asked $5$ individuals to tell for each of them whether they are ``near-perfect'', 
``perturbed but clearly identifiable'', ``hard to identify'', or ``rubbish'', and which digit they represent.
The results are that $96$ images were correctly identified; among them $63$ images were declared ``near-perfect'' by all individuals, 
with another $26$ including ``perturbed but clearly identifiable'', 
and $8$ were considered ``hard to identify'' by at least one individual yet correctly identified.
The remaining $4$ were ``rubbish'' or incorrectly identified.
It follows that against this generative model, we should require an image classifier to be at least $.89$-correct,
and even $.96$-correct to match human perception.

\subsection*{Algorithm}
To check whether a classifier is $\delta$-correct, the Monte Carlo integration method is a natural approach:
we sample $n$ random noises $x_1,\dots,x_n$,
and count for how many $x_i$'s we have that $\mC \circ \mG_c(x) = c$.
The central limit theorem states that the ratio of positives over $n$ converges 
to 
$\mathbb{P}_{x \sim \mathcal{N}(0,1)}(\mC \circ \mG_c(x) = c)$
as $\frac{1}{\sqrt{n}}$.
It follows that $n = 10^4$ samples gives a $10^{-2}$ precision on this number.

In practice, rather than sampling the random noises independently, 
we form (large) batches and leverage the tensor-based computation,
enabling efficient GPU computation.

\begin{figure*}[ht]
\centering
\includegraphics[scale=.9]{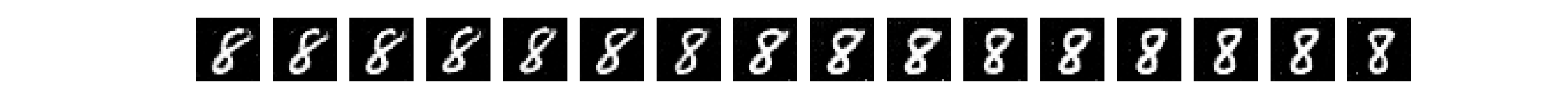} 
\caption{The random walk experiment}
\label{fig:random_walk}
\end{figure*}

\section{Global Robustness}
We introduce the notion of global robustness, which gives stronger guarantees than global correctness.
Indeed, it includes the notion of perturbations for images.

The usual notion of robustness, which we call here \textit{local robustness}, can be defined as follows.

\begin{definition}[Local Robustness]
We say that the image classifier $\mC$ is $\varepsilon$-robust around the image $y \in \R^d$ of category $c$ if
\[
\forall y',\ ||y - y'|| \le \varepsilon \implies \mC(y') = c.
\]
In words, all $\varepsilon$-perturbations of $y$ are correctly classified (by $\mC$).
\end{definition}
One important aspect in this definition is the choice of the norm for the perturbations (here we use the infinity norm).
We ignore this as it will not play a role in our definition of robustness.
A wealth of techniques have been developed for checking local robustness of neural networks, 
with state of the art tools being able to handle nets with thousands of neurons.

\subsection*{Assumptions}
Our definition of global robustness is supported by the two properties of generative models discussed above in the context of global correctness,
plus a third one:
\begin{enumerate}
	\item[3.] generative models produce perturbations of realistic images.
\end{enumerate}
To illustrate this we designed a second experiment called the \textit{random walk experiment}: 
we perform a random walk on the space of random noises while observing the ensued sequence of images produced by the generative model.
More specifically, we pick a random noise $x_0$, and define a sequence $(x_i)_{i \ge 0}$ of random noises with $x_{i+1}$ obtained from $x_i$
by adding a small random noise to $x_i$; this induces the sequence of images $(\mG(x_i))_{i \ge 0}$.
The result is best visualised in an animated GIF (see the Github repository), see also the first $16$ images in Figure~\ref{fig:random_walk}.
This supports the claim that images produced with similar random noises are (often) close to each other; 
in other words the generative model is (almost everywhere) continuous.

Our definition of global robustness is reminiscent of the \textit{provably approximately correct} learning framework developed by~\cite{Valiant84}.
It features two parameters.
The first parameter, $\delta$, quantifies the probability that a generative model produces a realistic image.
The second parameter, $\varepsilon$, measures the perturbations on the noise, 
which by the continuity property discussed above transfers to perturbations of the produced images.

\begin{definition}[Global Robustness]
Given for each $c \in C$ a generative model $\mG_c$ for images of category~$c$,
we say that the image classifier $\mC$ is $(\varepsilon,\delta)$-robust with respect to the generative models $(\mG_c)_{c \in C}$ 
if for each $c \in C$,
\[
\mathbb{P}_{x \sim \mathcal{N}(0,1)}(\forall x',\ ||x - x'|| \le \varepsilon \implies \mC \circ \mG_c(x') = c) \ge 1 - \delta.
\]
In words, the probability that for a noise $x$, all $\varepsilon$-perturbations of $x$ generate (using $\mG$) images 
correctly classified (by $\mC$) is at least $1 - \delta$.
\end{definition}

\subsection*{Algorithm}
To check whether a classifier is $(\varepsilon,\delta)$-robust, we extend the previous ideas using the Monte Carlo integration:
we sample $n$ random noises $x_1,\dots,x_n$,
and count for how many $x_i$'s the following property holds:
\[
\forall x,\ ||x_i - x|| \le \varepsilon \implies \mC \circ \mG_c(x) = c.
\]
The central limit theorem states that the ratio of positives over $n$ converges 
to 
\[
\mathbb{P}_{x \sim \mathcal{N}(0,1)}(\forall x',\ ||x - x'|| \le \varepsilon \implies \mC \circ \mG_c(x') = c)
\]
as $\frac{1}{\sqrt{n}}$.
As before, it follows that $n = 10^4$ samples gives a $10^{-2}$ precision on this number.

In other words, checking global robustness reduces to combining Monte Carlo integration with checking local robustness.

\section{Experiments}
The code for all experiments can be found on the Github repository 
\begin{center}
\url{https://github.com/mohitiitb/NeuralNetworkVerification_GlobalRobustness}.
\end{center}
All experiments are presented in Jupyter notebook format with pre-trained models to be easily reproduced.
Our experiments are all reproduced on the drop-in Fashion-MNIST dataset (\cite{XRV17}), obtaining similar results.

We report on experiments designed to assess the benefit of these two notions, 
whose common denominator is to go from a local property to a global one by composing with a generative model.

We first evaluate the global correctness of several image classifiers, 
showing that it provides a finer way of evaluating them than the usual test set.
We then turn to global robustness and show how the negation of robustness can be witnessed by realistic adversarial examples.

The second set of experiments addresses the fact that both global correctness and robustness notions depend on the choice of a generative model.
We show that this dependence can be made small, but that it can also be used for refining the correctness and robustness notions.

\subsection*{Choice of networks}
In all the experiments, our base case for image classifiers have $3$ hidden layers of increasing capacities:
the first one, referred to as ``small'', has layers with $(32,64,200)$ (number of nodes), 
``medium'' corresponds to $(64,128,256)$, and ``large'' to $(64,128,512)$.
The generative model are as described above, with $4$ hidden layers of respectively $256$, $512, 1024$, and $784$ nodes.

For each of these three architectures we either use the standard MNIST training set (6,000 images of each digit),
or an augmented training set (24,000 images), obtained by rotations, shear, and shifts.
The same distinction applies to GANs: the ``simple GAN'' uses the standard training set, 
and the ``augmented GAN'' the augmented training set.

Finally, we work with two networks obtained through robust training procedures.
The first one was proposed by~\cite{MadryMSTV18} for the MNIST Adversarial Example Challenge 
(the goal of the challenge was to find adversarial examples, see below), 
and the second one was defined by~\cite{PapernotMWJS15} through the process of defense distillation.

\subsection*{Evaluating Global Correctness}
\label{subsec:global_correctness}
We evaluated the global correctness of all the image classifiers mentioned above against simple and augmented GANs,
and reported the results in the table below.
The last column is the usual validation procedure, meaning the number of correct classification on the MNIST test set of 10,000 images.
They all perform very well, and close to perfectly (above $99\%$), against this metric, hence cannot be distinguished.
Yet the composition with a generative model reveals that their performance outside of the test set are actually different.
It is instructive to study the outliers for each image classifier, \textit{i.e.} the generated images which are incorrectly classified.
We refer to the Github repository for more experimental results along these lines.

\begin{figure*}
\centering
\begin{tabular}{c||c|c|c}
  \textbf{Classifier} & \textbf{simple GAN} & \textbf{augmented GAN} & \textbf{test set} \\
  \hline
  \multicolumn{4}{|c|}{Standard training set} \\
  \hline
  small & 98.89 & 92.82 & 99.79 \\
  medium & 99.15 & 93.16 & 99.76 \\
  large & 99.38 & 93.80 & 99.80 \\
  \hline
  \multicolumn{4}{|c|}{Augmented training set} \\
  \hline
  small & 97.84 & 95.2 & 99.90 \\
  medium & 99.11 & 96.53 & 99.86 \\
  large & 99.25 & 97.66 & 99.84 \\
  \hline
  \multicolumn{4}{|c|}{Robust training procedures} \\
  \hline
  \cite{MadryMSTV18} & 98.87 & 93.17 & 99.6 \\
  \cite{PapernotMWJS15} & 99.64 & 94.78 & 99.17
\end{tabular}
\end{figure*}

\subsection*{Finding Realistic Adversarial Examples}
\label{subsec:adversarial_examples}
Checking the global robustness of an image classifier is out of reach for state of the art verification tools.
Indeed, a single robustness check on a medium size net takes somewhere between dozens of seconds to a few minutes,
and to get a decent approximation we need to perform tens of thousands local robustness checks.
Hence with considerable computational efforts we could analyse one image classifier, 
but could not perform a wider comparison of different training procedures and influence on different aspects.
Thus our experiments focus on the negation of robustness, 
which is finding realistic adversarial examples, that we define now.

\begin{definition}[Realistic Adversarial Example]
An $\varepsilon$-realistic adversarial example for an image classifier $\mC$ with respect to a generative model $\mG$
is an image $\mG(x)$ such that there exists another image $\mG(x')$ with 
\[
||x - x'|| \le \varepsilon
\text{ and } 
\mC \circ \mG(x) \neq \mC \circ \mG(x')
\]
In words, $x$ and $x'$ are two $\varepsilon$-close random noises which generate images $\mG(x)$ and $\mG(x')$
that are classified differently by $\mC$.
\end{definition}
Note that a realistic adversarial example is not necessarily an adversarial example: 
the images $\mG(x)$ and $\mG(x')$ may differ by more than $\varepsilon$.
However, this is the assumption 3. discussed when defining global robustness, if $x$ and $x'$ are close,
then \textit{typically} $\mG(x)$ and $\mG(x')$ are two very resemblant images,
so the two notions are indeed close.

We introduce two algorithms for finding realistic adversarial examples, 
which are directly inspired by algorithms developed for finding adversarial examples.
The key difference is that realistic adversarial examples are searched by analysing the composed network $\mC \circ \mG$.

Let us consider two digits, for the sake of explanation, $3$ and $8$.
We have a generative model $\mG_8$ generating images of $8$ and an image classifier $\mC$.

The first algorithm is a \textit{black-box attack}, meaning that it does not have access to the inner structure of the networks
and it can only simulate them.
It consists in sampling random noises, and performing a local search for a few steps.
From a random noise $x$, we inspect the random noise $x + \delta$ for a few small random noises $\delta$,
and choose the random noise $x'$ maximising the score of $3$ by the net $\mC \circ \mG_8$, written 
$\mC \circ \mG_8(x_i)[3]$ in the pseudocode given in Algorithm~\ref{algo:black_box}.
The algorithm is repeatedly run until a realistic adversarial example is found.

\begin{algorithm}
 \KwData{A generative model $\mG_8$ and an image classifier $\mC$. A parameter $\varepsilon > 0$.}

$N_{\text{step}} \leftarrow 16$ (number of steps)

$N_{\text{dir}} \leftarrow 10$ (number of directions)

$x_0 \sim \mathcal{N}(0,1)$

\For{$i = 0$ to $N_{\text{step}} - 1$}{

	$s_{\text{max}} \leftarrow \mC \circ \mG_8(x_i)[3]$ (score of $3$)

	$x_{i+1} \leftarrow x_i$
	
	\For{$j = 0$ to $N_{\text{dir}} - 1$}{

		$\delta_j \sim \mathcal{N}(0,\frac{\varepsilon}{N_{\text{step}}})$

		$s \leftarrow \mC \circ \mG_8(x_i + \delta_j)[3]$

		\If{$s > s_{\text{max}}$}{
			$s_{\text{max}} \leftarrow s$

			$x_{i+1} \leftarrow x_i + \delta_j$
		}
	}

	\If{$\mC \circ \mG_8(x_0) \neq \mC \circ \mG_8(x_{i+1})$}{
		\Return{$x_0$} ($\varepsilon$-realistic adversarial example)
	}
}     

\caption{The black-box attack for the digits $3$ and $8$.}
\label{algo:black_box}
\end{algorithm}

The second algorithm is a \textit{white-box attack}, meaning that it uses the inner structure of the networks.
It is similar to the previous one, except that the local search is replaced by a gradient ascent to maximise 
the score of $3$ by the net $\mC \circ \mG_8$.
In other words, instead of choosing a direction at random, it follows the gradient to maximise the score.
It is reminiscent of the projected gradient descent (PGD) attack, but performed on the composed network.
The pseudocode is given in Algorithm~\ref{algo:white_box}.

\begin{algorithm}
 \KwData{A generative model $\mG_8$ and an image classifier $\mC$. A parameter $\varepsilon > 0$.}

$N_{\text{step}} \leftarrow 16$ (number of steps)

$\alpha \leftarrow \frac{\varepsilon}{N_{\text{step}}}$ (step)

$x_0 \sim \mathcal{N}(0,1)$

\For{$i = 0$ to $N_{\text{step}} - 1$}{

	$x_{i+1} \leftarrow x_i - \alpha \cdot \text{Grad}_{\mC \circ \mG_8}(x_i)[3]$

	\If{$\mC \circ \mG_8(x_0) \neq \mC \circ \mG_8(x_{i+1})$}{
		\Return{$x_0$} ($\varepsilon$-realistic adversarial example)
	}
}     

\caption{The white-box attack for the digits $3$ and $8$.}
\label{algo:white_box}
\end{algorithm}

\begin{figure*}
\centering
\begin{minipage}{.5\textwidth}
  \centering
  \includegraphics[width=.4\linewidth]{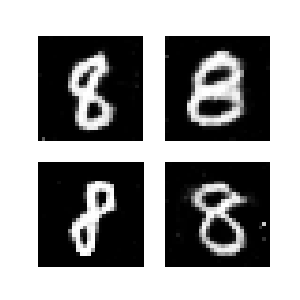}
\end{minipage}%
\begin{minipage}{.5\textwidth}
  \centering
  \includegraphics[width=.4\linewidth]{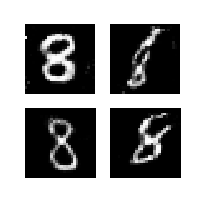}
\end{minipage}
\caption{Examples of realistic adversarial examples. 
On the left hand side, against the smallest net, 
and on the right hand side, against~\cite{MadryMSTV18}}
\label{fig:adversarial_examples}
\end{figure*}

Both attacks successfully find realistic adversarial examples within less than a minute.
The adjective ``realistic'', which is subjective, is justified as follows: 
most attacks constructing adversarial examples create unrealistic images by adding noise or modifying pixels,
while with our definition the realistic adversarial examples are images produced by the generative model,
hence potentially more realistic. 
See Figure~\ref{fig:adversarial_examples} for some examples.

\subsection*{On the Dependence on the Generative Model}
\label{subsec:dependence_model}
Both global correctness and robustness notions are defined with respect to a generative model.
This raises a question: how much does it depend on the choice of the generative model?

To answer this question we trained two GANs using the exact same training procedure but with two disjoint training sets,
and used the two GANs to evaluate several image classifiers.
The outcome is that the two GANs yield sensibly the same results against all image classifiers. 
This suggests that the global correctness indeed does not depend dramatically on the choice of the generative model,
provided that it is reasonably good and well-trained.
We refer to the Github repository for a complete exposition of the results.

Since the training set of the MNIST dataset contains 6,000 images of each digit,
splitting it in two would not yield two large enough training sets. 
Hence we used the extended MNIST (EMNIST) dataset~\cite{CohenATS17},
which provided us with (roughly) 34,000 images of each digit, hence two disjoint datasets of about 17,000 images.

\subsection*{On the Influence of Data Augmentation}
\label{subsec:data_augmentation}
Data augmentation is a classical technique for increasing the size of a training set,
it consists in creating new training data by applying a set of mild transformations to the existing training set.
In the case of digit images, common transformations include rotations, shear, and shifts.

Unsurprisingly, crossing the two training sets, \textit{e.g.} using the standard training set for the image classifier
and an augmented one for the generative model yields worse results than when using the same training set.
More interestingly, the robust networks~\cite{MadryMSTV18,PapernotMWJS15}, 
which are trained using an improved procedure but based on the standard training set, 
perform well against generative models trained on the augmented training set.
In other words, one outcome of the improved training procedure is to better capture the natural image transformations,
even if they were never used in training.

\section{Conclusions}
We defined two notions: global correctness and global robustness, based on generative models, 
aiming at quantifying the usability of an image classifier.
We performed some experiments on the MNIST dataset to understand the merits and limits of our definitions.
An important challenge lies ahead: to make the verification of global robustness doable in a reasonable amount of time and computational effort.


\bibliographystyle{plainnat}
\bibliography{bib}

\end{document}